\documentclass[lettersize,journal]{IEEEtran}
\usepackage{amsmath,amsfonts}
\usepackage{algorithmic}
\usepackage{algorithm}
\usepackage{array}
\usepackage[caption=false,font=normalsize,labelfont=sf,textfont=sf]{subfig}
\usepackage{textcomp}
\usepackage{stfloats}
\usepackage{url}
\usepackage{verbatim}
\usepackage{graphicx}
\usepackage{newunicodechar}
\newunicodechar{ }{~}
\usepackage{cite}
\usepackage{gensymb}

\begin{document}

\title{Replaceable Bit-based Gripper for Picking Cluttered Food Items}

\author{Prashant Kumar, Yukiyasu Domae, Weiwei Wan, and Kensuke Harada}



\maketitle

\begin{abstract}
 The food packaging industry goes through changes in food items and their weights quite rapidly. These items range
 from easy-to-pick, single-piece food items to flexible, long and cluttered ones. We propose a replaceable bit-based gripper system to
 tackle the challenge of weight-based handling of cluttered food items. The gripper features specialized food attachments(bits) that enhance its grasping capabilities,
 and a belt replacement system allows switching between different food items during packaging operations. It offers a wide range
 of control options, enabling it to grasp and drop specific weights of granular, cluttered, and entangled foods. We specifically designed
 bits for two flexible food items that differ in shape: ikura(salmon roe) and spaghetti. They represent the challenging categories of sticky,
 granular food and long, sticky, cluttered food, respectively. 
 

The gripper successfully picked up both spaghetti and ikura and demonstrated weight-specific dropping of these items with an accuracy over 80\% and 95\% respectively. The gripper system also exhibited quick switching between different bits, leading to the handling of a large range of food items.
 
\end{abstract}


\section{Introduction}
\IEEEPARstart{A}{utomated} food handling systems have gained increasing interest due to the rising demand for consistent, high-throughput food packaging. However, the automation of bento box preparation remains a significant challenge\cite{wangsoft}. Bento boxes are traditional Japanese single-portion meals that include a carefully arranged assortment of diverse food items—such as rice, fish, vegetables, and side dishes, each with different textures, shapes, and mechanical properties.
Three key factors that make bento box automation particularly difficult:
(1) A significant portion of bento box components, such as ikura (salmon roe), spaghetti, and shredded vegetables, are fragile, deformable, slippery, or tend to get tangled, making them hard to pick up and place reliably;
(2) precise weight control of challenging food items is essential to ensure portion consistency, especially for commercial packaging and calorie regulation; and
(3) the wide range of food types requires grippers to adapt quickly to different physical characteristics, and the target weight of each food type may vary across different boxes, further complicating the handling process.\\

There has been a wide body of work on food handling and food box packaging. One popular direction has been the development of soft grippers designed to gently pick up food items with minimal contact force\cite{kanegae2020easily,kumar2025temperature,wang2020dual,wang2021circular}.  However, these approaches are generally limited to handling single-piece food items (e.g., berries, fried chicken) placed separately in controlled environments. Further research has extended this capability to cluttered scenes, using vision and learning-based methods to pick out individual items\cite{ummadisingu2022cluttered}. Ummadisingu et al. used a machine learning approach to identify and grasp isolated food items from complex, cluttered settings. While these strategies are effective for discrete, well-separated items, such food types make up only a small fraction of bento box contents. The majority of bento components are granular, entangled, or deformable—such as shredded cabbage, herbs, or noodles—posing unique manipulation challenges. Several researchers have addressed this by designing adaptive grippers capable of conforming to irregular, granular food items. \cite{wrappinggripper,wang2021scooping,ray2020robotic}. These approaches mainly tried to leverage the adaptability and softness of the gripper setup to be able to grasp granular food items such as chopped cabbage, herbs etc. 
However, they generally lack the ability to control the amount of food picked or dropped, especially in terms of achieving a target weight.

\begin{figure}[t]
\centering
\includegraphics[width=3.5in]{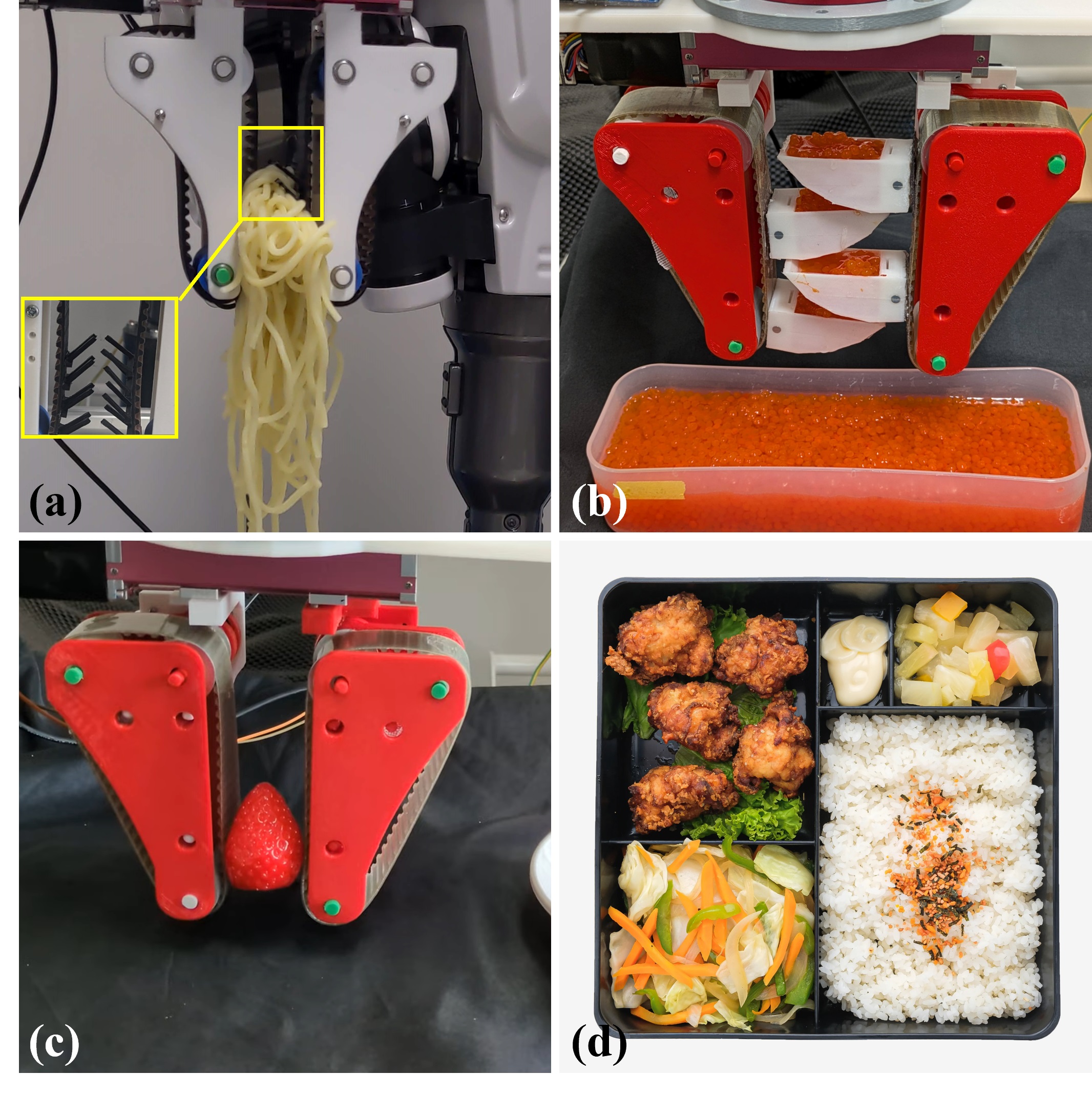}
\caption{Using food specific bits to pick up food (a) Spaghetti being picked up. (b) Ikura pick up. (c) Strawberry pick up. (d) Standard Japanese lunch box.}
\label{fig_1}
\end{figure}

Another popular approach has been to train a parallel jaw gripper setup using deep learning to pick up food with its accurate weight\cite{takahashi2021target,takahashi2021uncertainty}. 
While effective in limited settings, these methods are not easily scalable to the diverse range of foods typically found in bento boxes. The requirement for large training datasets and long training times limits their practical use in dynamic food packaging environments.



To meet the unique requirements of bento box automation, a solution must simultaneously (1) perform reliable grasps on challenging food types, (2) drop food with controlled and accurate weight, and (3) adapt quickly across a wide range of ingredients. We propose a hardware-centric robotic gripper system that addresses all three of these challenges.

Inspired by a Panasonic gripper design that uses a movable belt surface for object manipulation \cite{panasonic2024foodrobots}, our gripper integrates food-specific attachment (hereby referred to as bits) onto an active belt surface. This hybrid design allows the gripper to pick up and release complex, deformable, or entangled food items—such as spaghetti or salmon roe—in a controlled and weight-specific manner. In addition, we developed an automatic tool-changing mechanism that enables rapid switching between different food-specific attachments. The gripper’s fully passive mechanism enables effortless tool switching, ensuring smooth transitions between food categories without active intervention.

The contributions of the paper can be concluded as below:
1. Bit based gripper for effective handling of challenging food items.\\
2. Weight specific drop of food items, achieving over 87 percent accuracy.\\
3. A passive and quick tool changing mechanism for switching between food-specific gripper modules.\\
4. A unified system capable of packaging a wide range of challenging, deformable food items in targeted quantities for automated bento box preparation.

\begin{figure*}[t]
\centering
\includegraphics[width = 0.8\textwidth]{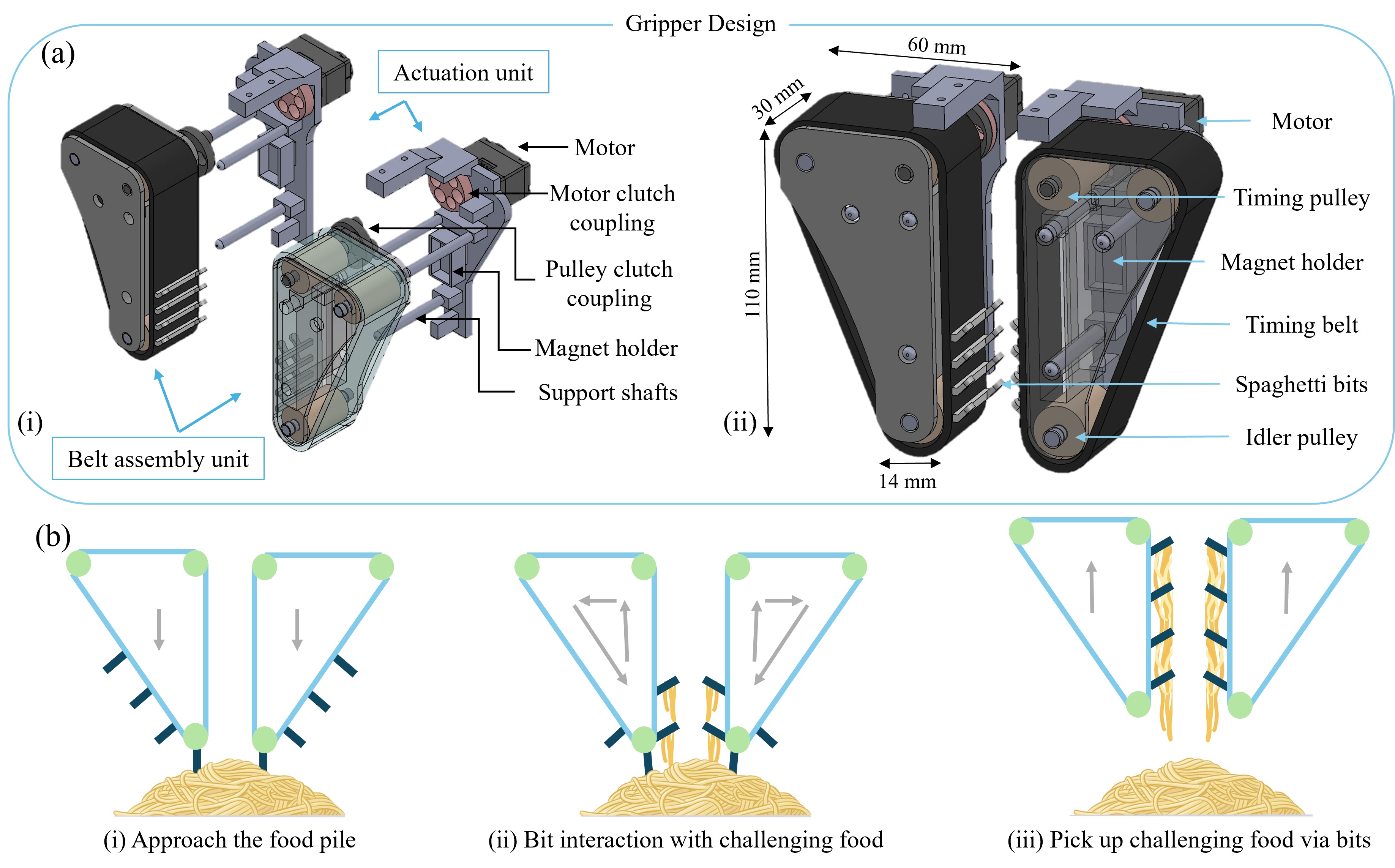}
\caption{(a).Gripper design (i) Two different parts of the gripper setup: Actuation unit and the Belt assembly unit. (ii) Components of the belt assembly unit. (b).Working of the gripper (i) Gripper approaches the food pile with bits pointitng towards the food for easier entry. (ii) The motors turn the belt and the bits move upwards, interacting and picking up the challenging food item. (iii) The bits ensure that the food is securely grasped and remains in place as the gripper moves away.}
\label{mss}
\end{figure*}

\section{Related Research} 

Food box packaging research mainly comprises research on picking up single food items, picking of granular and challenging food items, and weight-based food pickup, which together allows a system to pick up the required weight of a food item and place it in the targeted bento box. We have summarized these works in the following two subsections.
\subsection{Robotic food handling}
Attempts at robotic food manipulation date back decades. The softness and delicate nature of food has always made the pickup challenging\cite{qiu2023evaluation}. 
A significant part of food handling research has explored picking up single piece food items. One of the most popular approaches is to utilize the adaptability of soft grippers to handle delicate food items\cite{kanegae2020easily,kumar2025temperature,wang2020dual,wang2021circular,wangsoft,wang2017prestressed}. \cite{kanegae2020easily}Kanegae et al. developed a soft gripper in the shape of a circular shell to grasp objects of various shapes. Kumar et al.\cite{kumar2025temperature} developed a multi-modal gripper that could pick up delicate items like a tofu cube by coiling the soft gripper around it. 
Another popular approach is to utilize tactile sensing to observe the rigidity/softness of food and adjust the grasping force accordingly to grasp without damaging the food\cite{Ishikawa_tactile,Yagawa_tactile} or manipulation policy-centric food acquisition policies such as scooping\cite{grannen2022scooping}. Ummadisingu et al.\cite{ummadisingu2022cluttered} took this a step ahead and used a training based approach to identify and grasp single-piece food items from cluttered settings.

The developed approaches were limited to single-piece food items. Challenging categories such as long, entangled food, sticky, or slippery granular food were still unexplored. 
Several approaches were used to grasp granular entangled food items\cite{wang2021scooping,schenck2017learning,kuriyama2019wrapping,ray2020robotic,tai2023scone}. Wang. et al.\cite{wang2021scooping} tackled this challenge by developing a scooping gripper capable of scooping up finely chopped cabbage pieces. Kuriyama et al.\cite{kuriyama2019wrapping} developed a wrapping gripper that could grasp small granular items such as corn, beans. and even powder.  Ray et al.\cite{ray2020robotic} used a FGE(Fast Graspability Evaluation)\cite{domae2014fast} based approach to evaluate the best pickup point for individual herbs from a pile of entangled herbs. Although these approaches could handle small granular food items, they were largely limited to fine-grained materials and only mildly entangled food types. But more importantly, these approaches lacked the key factor of picking up these food items on the basis of their weights.

\begin{figure*}[t]
\centering
\includegraphics[width = 0.8\textwidth]{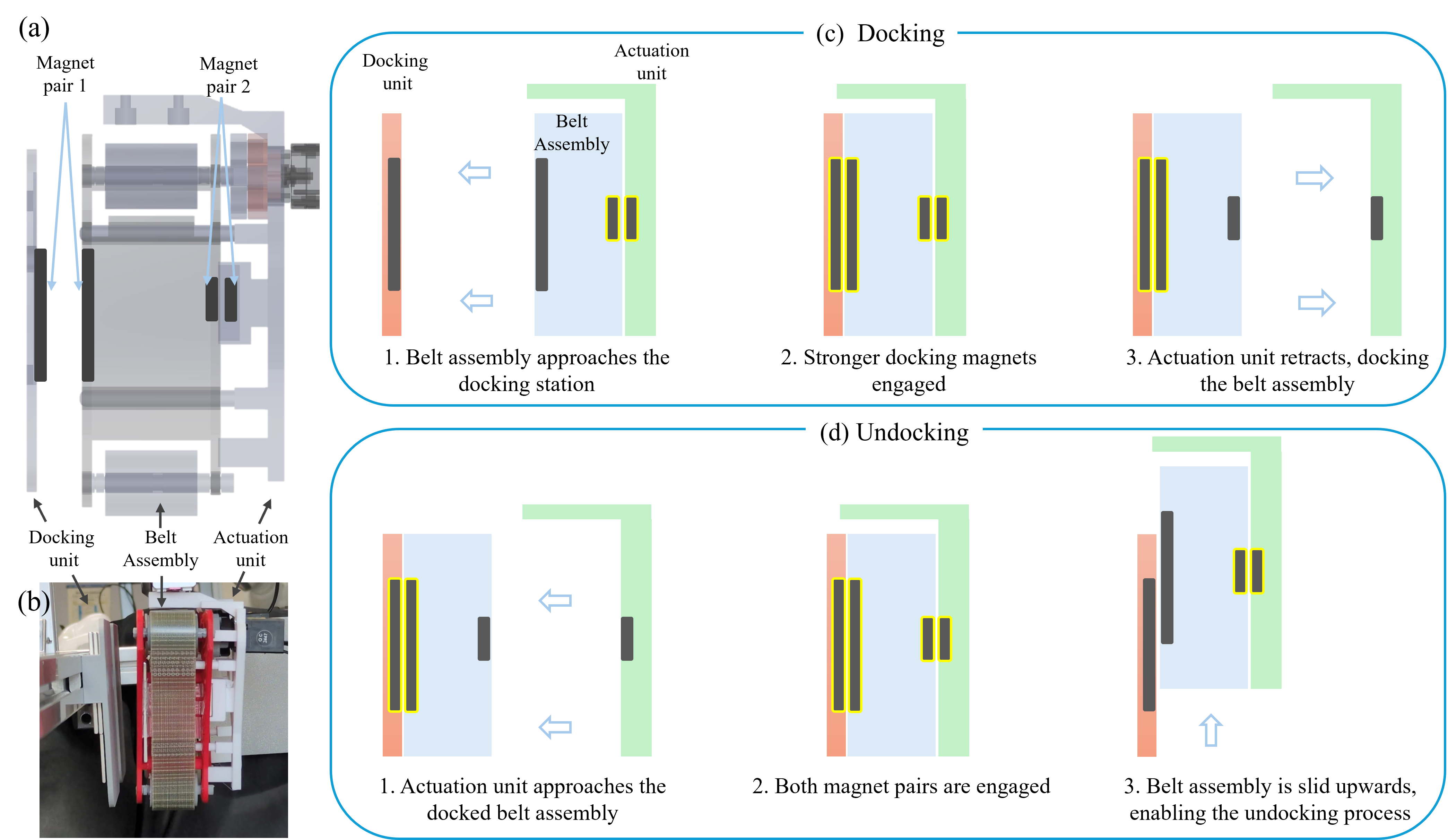}
\caption{Working of the belt assembly changer mechanism. (a) CAD of three parts of the belt assembly changing setup. (b) Image of the actual setup. (c) Docking process i.e. removing the belt assembly from the gripper and placing it at the docking station. (d) Undocking process i.e transferring the belt assembly from the docking station to the robotic arm gripper setup.}
\label{mss}
\end{figure*}                                                                             

\subsection{Weight-based food pickup}
Research on weight-based food picking remains largely unexplored, with only a few successful cases reported. This highlights the need for effective and practical solutions to address this critical challenge.
Takahashi et al.\cite{takahashi2021target} used a deep learning approach to create a system capable of picking up the required weight of rice just with an RGB-D input of the rice container. Takahashi et al.\cite{takahashi2021uncertainty} advanced this research by adding pre and post grasping steps to pick up short but granular food item like cabbage, beans. These aproaches provided good results but requires high amount of data for every single food item and with the large variety of bento boxes, this could be troublesome to implement. \\

While prior work has made significant advances in food grasping, most systems are specialized either for isolated food items or lack precise weight control when handling deformable or granular food. Furthermore, learning-based systems require significant training time and lack flexibility. Our work addresses these limitations through a modular, hardware-centric design that can reliably pick up a wide range of challenging food items with accurate weight control and rapid tool-switching.



\section{Replaceable bit-based gripper}

In this section, we introduce our developed Bit-based gripper system that can pick up challenging food items and drop specific weights of it. 
The combination of food-specific attachments on a moving surface allows the gripper to pick up these food items. It can also cater to a large range of food items, making it suitable to handle the variations of challenging food items commonly present in food boxes. This gripper solves the challenge of automating food box packaging, which involves handling a large range of food items and precisely weight-based dropping of these items in various boxes.

 The developed gripper has two major features and functions:\\
 1. The bit based gripper ensures the picking up of challenging food items \\
 2. Belt assembly changing mechanism increases the range of food that the gripper can handle.
 \\They have been explained in the following two subsections:

\subsection{Bit Based Gripper}
\subsubsection{Design}
The developed gripper consists of two parts: a main actuation unit and a food-specific belt assembly(Fig.2).
The main actuation unit includes a servo motor and the mounting interface with the robotic arm. It serves as the driving mechanism for the belt assembly. Motion is transmitted via a pin-coupling clutch shaft(refer Fig.2), enabling quick and secure connection to various belt assembly modules.

The belt assembly is designed as a modular unit tailored for specific food categories. It consists of a timing belt wrapped around three pulleys—two located at the top and one at the bottom. This triangular configuration forms a narrow, tapered structure at the bottom, which facilitates minimal disturbance and smooth entry into food bowls or containers.

Most importantly, the timing belt itself is embedded with food-specific attachments, referred to as bits. Their role is to increase the gripper’s ability to grasp, pick up, and drop challenging food items, such as entangled, sticky, or granular materials. The motion of the belt, controlled by the actuation unit, drives these bits to effectively pick up and release such food items. The shape and structure of the bits are inspired by real-world tools commonly used to interact with objects of similar textures and behaviors.



\begin{figure}[t]
\centering
\includegraphics[width=3.4in]{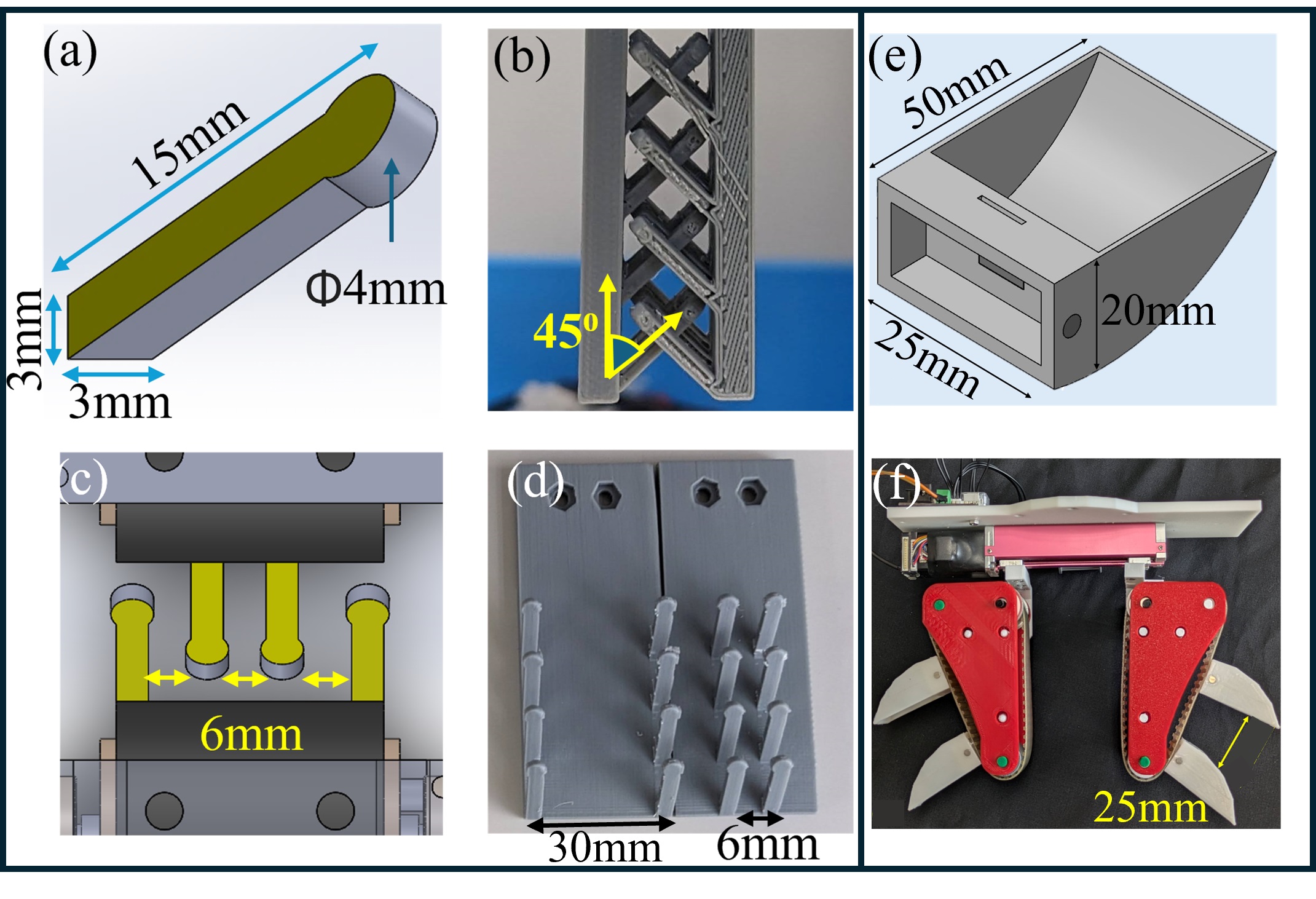}
\caption{Bit parameters for spaghetti(a)Individual bit inspired
from a spaghetti spoon. (b)Bits positioned at 45\degree w.r.t the
belt. (c)Bits arrangement from the top view of the gripper. (d)Bit density(low): Individual plates displaying 6mm distance between each bit. Gripper width is 30mm. (e)Bit parameters for ikura. (f)Distance of 25mm between each ikura bit.}
\label{fig_1}
\end{figure}


As mentioned earlier, we evaluated the gripper’s performance
 on two food item categories. We chose these categories as they present significant interactive challenges and cover a large range of commonly occurring and challenging food items in food boxes.
 Following is the information about the
 bit selection for these food items:\\
 
1.1 Long, entangled food:
 To represent the category of long, entangled foods commonly found in Japanese lunch boxes, we selected spaghetti as a test item. The design of the spaghetti-specific bit was inspired by the spokes of a traditional spaghetti spoon\cite{mot_pasta_spoon}, which are optimized for handling slippery and tangled strands.
The length of the bit was chosen to match that of a standard spaghetti spoon spoke, ensuring sufficient engagement with the strands. We set the bit angle at 45°, neutral angle for both picking and dropping(Fig. 4(b)). 



To determine optimal bit parameters, we conducted experimental evaluations with static gripper setups using different combinations of gripper width and bit density. Each setup performed the same pickup and drop motion on real spaghetti under identical conditions (10 trials each). We tested three gripper widths—20 mm, 30 mm, and 40 mm—and found that width had little effect on dropped weight. We selected 30 mm as it was compact enough for small food bowls while offering enough area to try different bit densities. We tested three bit densities: ~2 bits/cm² (high), ~1.5 bits/cm² (medium), and ~1 bit/cm² (low), as shown in Table 1. 

The goal of this evaluation was to identify the setup that could pick up and drop spaghetti effectively, with no accidental drops during transfer and minimal strand damage. We measured three outcomes: the weight successfully dropped in the target bowl, the number of strands accidentally dropped in transit, and the number of strands damaged by the bits (Table 1).
 The following is the score system:
 \[
\resizebox{0.49\textwidth}{!}{$
\text{Score}_{\text{spaghetti}} = w_1 \cdot \text{Dropped weight} - w_2 \cdot \text{Accidental drops} -
\\ w_3 \cdot \text{Damaged strands}
$}
\]

\begin{table}[thbp]
\caption{Spaghetti bit selection experiment parameters}
\centering
\resizebox{0.51\textwidth}{!}{
\begin{tabular}{|c|c|c|c|c|c|}
\hline
\textbf{\shortstack{Gripper\\width (mm)}} &  
\textbf{\shortstack{Bit density\\(bits/$cm^2$)}} & 
\textbf{\shortstack{Dropped\\weight (g)}} & 
\textbf{\shortstack{Accidental\\drops(strands)}} & 
\textbf{\shortstack{Spaghetti strands\\damaged}} & 
\textbf{\shortstack{Performance\\score}} \\
\hline
20 & High ($~2/cm^2$) & 42 & 1  & 4.8 & 31.4 \\
\hline
30 & High ($~2/cm^2$) & 39.3 & 0.5 & 5.83 & 27.14 \\
\hline
40 & High ($~2/cm^2$) & 45.2 & 0.6 & 6.8 & 31 \\
\hline
30 & Med. ($~1.5/cm^2$) & 41 & 0.14 & 2.28 & 36.3 \\
\hline
30 & Low ($~1/cm^2$) & 48.42 & 0.8 & 0.14 & 47.34 \\
\hline
\end{tabular}
}
\label{spaghetti bit experiment}
\end{table}

\begin{table}[htbp]
\caption{Ikura bit selection experiment parameters}
\centering
\resizebox{0.5\textwidth}{!}{%
\begin{tabular}{|c|c|c|c|c|}
\hline
\textbf{\shortstack{Ikura bit's\\scoop's curve}} &  
\textbf{\shortstack{Dropped\\ weight (g)}} & 
\textbf{\shortstack{No. of remaining\\ ikura in bit}} & 
\textbf{\shortstack{Ease of drop\\(drop angle)}} & 
\textbf{\shortstack{Performance\\ score}} \\
\hline
Conical & 7.401 & 2.6 & 3.62 (115\degree) & 9.7 \\
\hline
Elliptical & 12.28 & 5.8 & 3.3 (122\degree) & 12.68 \\
\hline
Circular & 11.206 & 3.5 & 4.3 (103\degree) & 13.756 \\
\hline
\end{tabular}%
}
\end{table}

where the weights $w_1=1$, $w_2=1$ and $w_3=2$, as we highly value not damaging the spaghetti. The 30 mm wide gripper with low bit density achieved the highest performance score, combining a high dropped weight with wide spacing between bits that enabled safe, damage-free handling.\\

1.2 Granular, slippery food: To represent granular and slippery food items in Japanese lunch boxes, we selected ikura (salmon roe) as a test item. The bit design for handling ikura was inspired by bucket-style mechanisms used in gravel conveyor systems and excavators\cite{kim20212d}, which are effective at scooping and transporting small and loose materials.
\\
The width and length were each set to 25 mm and 50mm, matching the width of the existing belt system(Fig.4e). Three different scoop curve profiles were tried, conical, elliptical and circular. We performed an evaluation test where these three setups picked up and dropped ikura(10 trials). The ease and effectiveness of ikura pickup and drop were evaluated and the performance scores were awarded. Once again, the score was evaluated by:
\[
\resizebox{0.49\textwidth}{!}{$
\text{Score}_{\text{ikura}} = w_1 \cdot \text{Dropped weight} - w_2 \cdot \text{Remaining ikura} +
\\ w_3 \cdot \text{Ease of drop}
$}
\]

where the weights are $w_1,w_3=1$ and $w_2=0.5$, as the amount picked up and the ease of drop are the most important requirements.
The circular profile achieved the highest score, demonstrating an effective balance between high pickup weight and smooth release with minimal tilt angle. Hence, a bucket like shape of width 25mm, length 50mm and a circular curve profile was chosen to handle the category of granular and slippery food.


1.3 Single piece food item: Fried chicken was chosen to represent single-piece foods, a common focus in prior studies\cite{ummadisingu2022cluttered,nishina2020model}. While less challenging than granular or entangled items, they remain essential in food packaging .Our gripper handled them using only the base configuration—a moving belt in a parallel jaw setup—without food-specific attachments (Fig. 1c), showcasing the gripper’s ability to adapt across a range of food types, from the most challenging to the relatively simple.


We also performed picking on other food items from these categories(entangled foods: diced carrots, bean sprouts, shredded cabbage, thick udon noodles. Granular foods: corn, peas, peanuts, red beans, etc) to showcase the effectiveness of handling and the ease of implementation over a large range of food (Supplementary Video).
It is to be noted that the overall concept being presented in the paper is attaching food-specific bits on a moving belt for controlled food picking and dropping. This concept is independent of any specific bit and can be expanded and implemented to any food category with ease.\\

\subsubsection{Working}
The gripper begins operation from its home position, where the food-specific bits are positioned outside the parallel jaw closing zone (Fig. 2b(i)). Initially, the gripper's jaws are closed using a linear actuator called seed actuator to guide the food from the sides toward the center. It is to be noted that the food pickup will work without this horizontal closing too but this step ensures a longer use time as the food is constantly collected from the sides to the middle.

The motor rotation turns the belt, which moves the food-specific bits upwards and picks up/lifts food from the source bowl. The picked-up food gets spread out in a segregated manner over the bits. This segregated state of food allows the gripper system to perform controlled dropping in the end. 


The highly entangled and long nature of spaghetti makes it difficult to pick up a specific weight, so instead, the gripper tends to pick up a weight up to its maximum limit and later on drops the required weight of the food.
 \\
To deposit food into designated containers, the gripper positions itself above the target bowl and rotates the belt in small, controlled increments (typically 20° per step). A digital weighing scale placed beneath the container provides real-time feedback, allowing the system to precisely monitor the weight of the dropped food. 

The food-specific bits on a moving belt play a crucial role in enabling this level of control. The food bits ensure a secure grip on challenging food items, while their segregated arrangement along the belt distributes the picked-up food across multiple compartments. This spatial separation allows the gripper to release food incrementally, providing fine-grained weight control that is difficult to achieve with conventional grippers. The gripper drops the target weight of food in the designated container and then proceeds to continue this for another container.\\

\subsection{Belt Changing mechanism}


As mentioned earlier, the gripper can accurately pick up and drop small quantities of challenging food items. However, since food packaging industries deal with not only different weights but also a large range of food items, the gripper was also designed with a passive, quick belt assembly changing feature. This enables switching between tools for different food items in under forty seconds. The feature not only supports handling diverse items but also prevents cross-contamination, preserving each item’s taste and freshness.
\subsubsection{Design}
A key design consideration for the belt assembly changer was to ensure that it remained actuator-independent, quick to operate, and simple to control. These features were essential to make the system easily scalable, allowing additional belt assemblies or tools to be integrated without increasing complexity or cost.

The first component of the changer is a docking station (Fig. 3), which contains an embedded 50 × 10 × 3 mm, 30N magnet. A corresponding magnet of the same dimensions is embedded on the inner face of the belt assembly (Fig. 3B). These long magnets provide a strong and reliable magnetic connection, while also enabling self-alignment in the plane of the docking surface during attachment. This design ensures fast, secure, and repeatable connections without the need for active alignment mechanisms.

The main actuation unit (Fig. 2) is designed to be compatible with belt assemblies of varying sizes and attachment types, provided the position of mounting holes remains consistent. The actuation unit includes a Dynamixel XL330-M288-T servo motor, three support shafts, and a 20 X 10 X 2mm, 10N strong alignment magnet. A corresponding magnet is embedded in the belt assembly to align and stabilize it during attachment. The three shafts provide mechanical alignment, while the magnet prevents lateral shifts, ensuring a stable connection during the gripper's operations.

Importantly, the belt assembly itself is entirely passive, consisting of pulleys, rollers, and the timing belt (as described in Section 3.A). This eliminates the need for additional actuators, keeping the system lightweight, cost-effective, easy to maintain, and scalable.

\begin{figure}[t]
\centering
\includegraphics[width=2.8in]{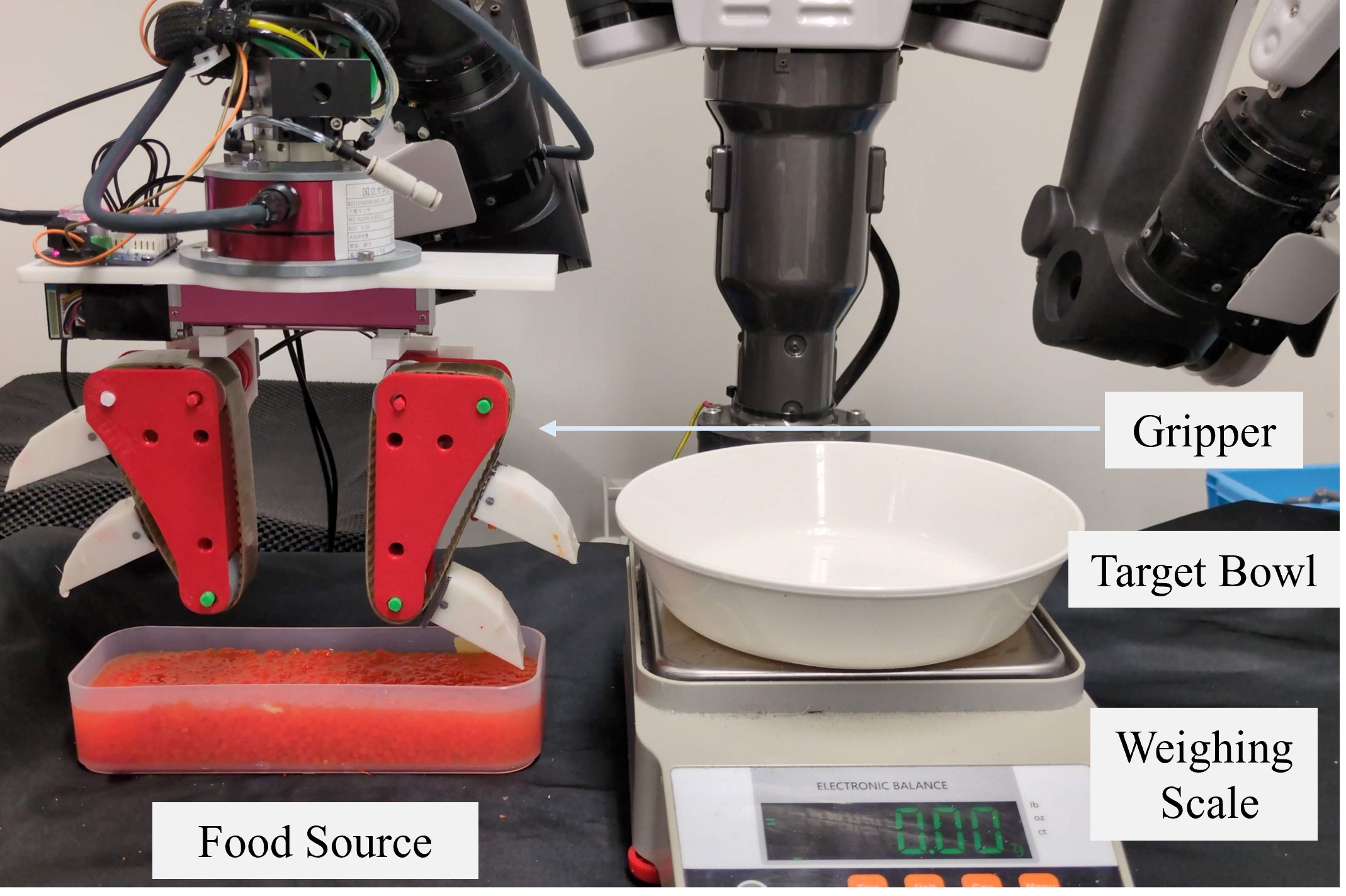}
\caption{Setup for weight-based drop experiment: Salmon roe(Ikura) and spaghetti(not in image) being picked up and dropped according to required weight. Nextage open arm and a digital weighing scale being employed.}
\label{fig_1}
\end{figure}


\subsubsection{Working}
Once the need to handle a different food item emerges, the robotic arm initiates the belt assembly change sequence. First, it aligns the current belt assembly co-axially with the docking station and proceeds to slowly moves towards the dock(Fig. 3.c.1). As the belt assembly approaches the dock, the magnetic attraction between the long embedded magnets in both components becomes effective. These magnets automatically pull and align the belt assembly into position(Fig. 3.c.2). The magnets in the docking station are deliberately made stronger, ensuring that the belt assembly remains attached to the dock when the actuation unit retracts (Fig. 3.c.3). This completes the docking or disengaging process of the current belt assembly.

To engage a new belt assembly, the robotic arm aligns the actuation unit coaxially with a different docking station that holds the desired food-specific assembly. It then advances the actuation unit toward the new assembly until the three alignment shafts fully insert into their respective holes(Fig. 3.d.1). Simultaneously, small internal magnets engage, securing the connection(Fig. 3.d.2).

Throughout this step, the system continuously monitors force sensor data to ensure that insertion forces remain within safe limits. Once the connection is established, the robotic arm slides the new belt assembly up at a controlled speed. During this lifting motion, the sliding action causes a reduction in magnetic force—as the magnets shift into a frictional, low-strength state, allowing the actuation unit to easily separate the belt assembly from the dock (Fig. 3.d.3). With the new assembly attached, the robotic system is now ready to resume operation with food-specific bits designed for the next item.

\begin{figure*}[htbp]
    \centering
    \begin{minipage}[t]{0.48\textwidth}
        \centering
        \includegraphics[width=\linewidth]{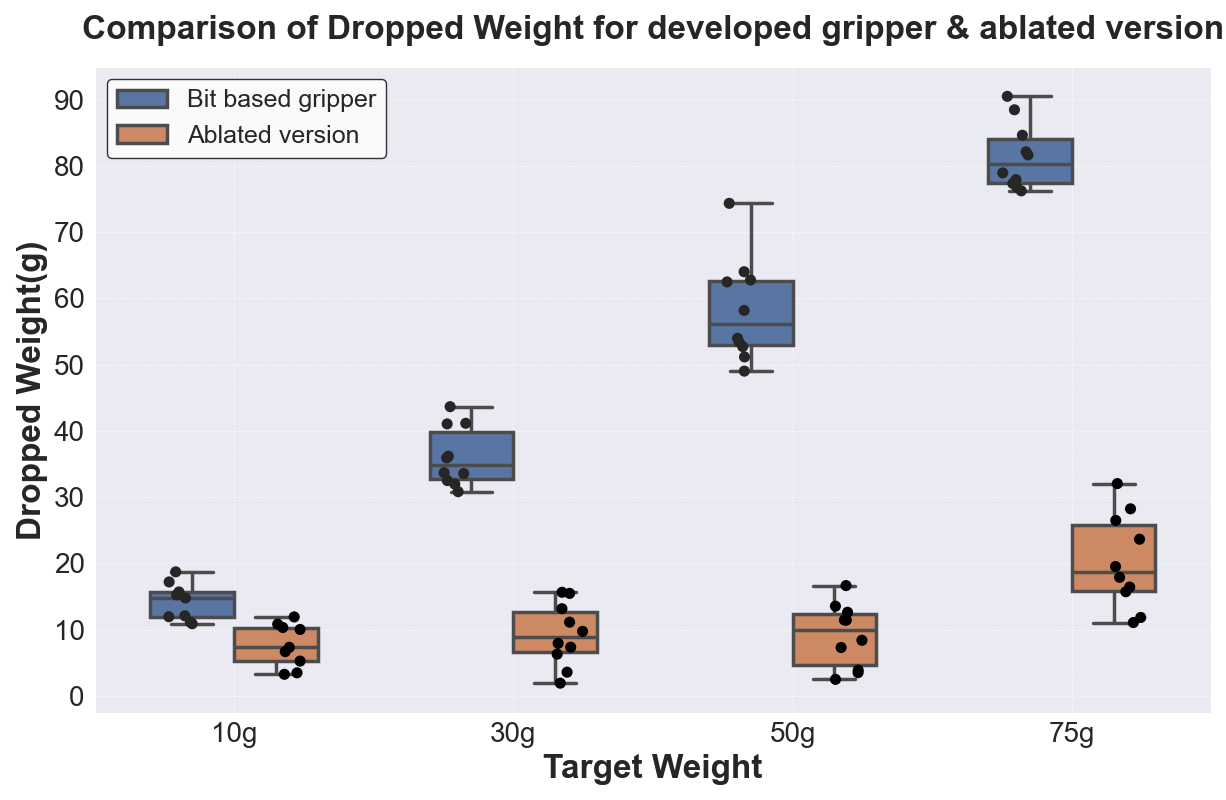}
        \caption{ Comparison of weight-based drop accuracy between the developed gripper and its ablated version for handling spaghetti }
        \label{fig:fig1}
    \end{minipage}
    \hfill
    \begin{minipage}[t]{0.48\textwidth}
        \centering
        \includegraphics[width=\linewidth]{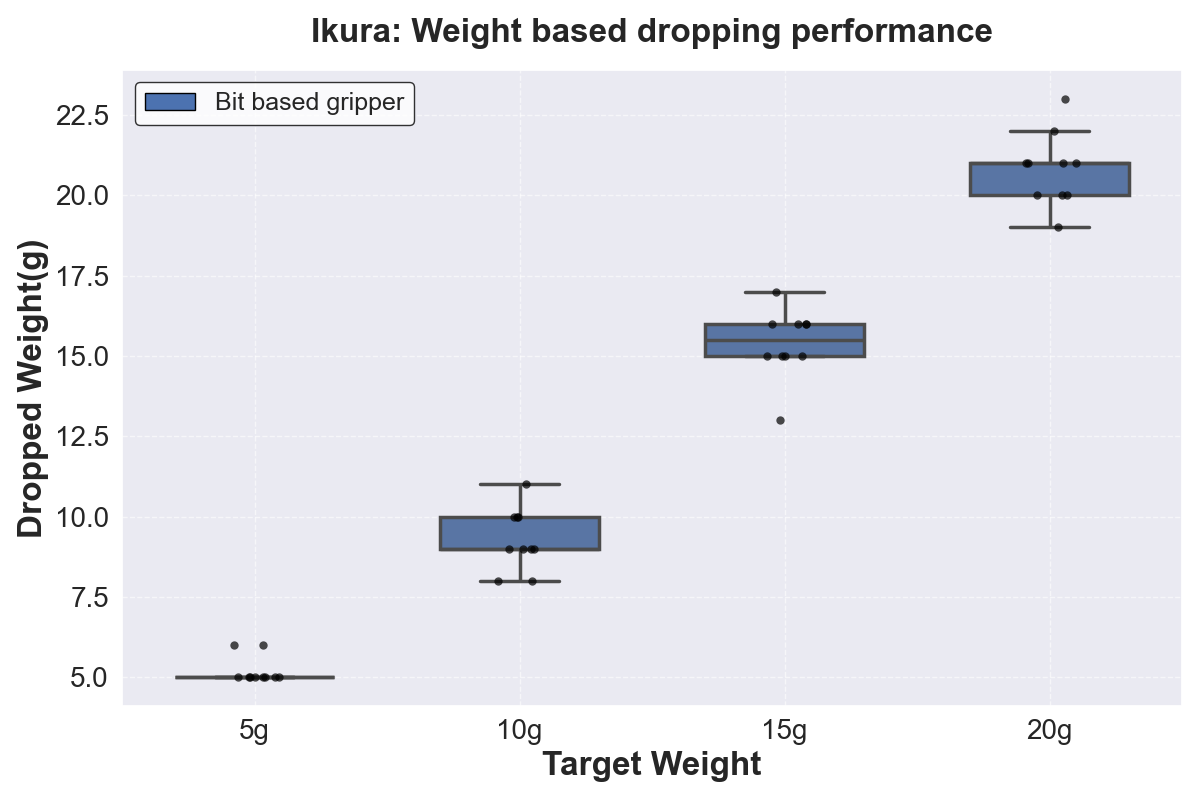}
        \caption{Weight-based drop accuracy of the gripper for handling ikura}
        \label{Fig6}
    \end{minipage}
\end{figure*}

 \begin{table*}[htbp]
\caption{Weight-based dropping performance: Spaghetti}
\centering
\resizebox{0.99\textwidth}{!}{%
\begin{tabular}{|c|c|c|c|c|c|c|}
\hline
\textbf{Target Weight(g)} & \textbf{Average dropped weight(g)} & \textbf{Accuracy(\%)} & \textbf{Standard Deviation} & \textbf{Ablated: dropped weight(g)} & \textbf{Ablated: accuracy\%} & \textbf{Ablated SD} \\
\hline
10 & 14.18 & 59\% & 2.81 & 7.67 & 67\% & 3.76 \\
\hline
30 & 36.185 & 79.4\% & 4.21 & 9.22 & 31\% & 4.137 \\
\hline
50 & 59.558 & 81\% &7.73  &9.127  &19.3\%  & 5.03 \\
\hline
 75&80.894  & 92.2\% & 5.02 &20.79  &28\%  &7.069  \\
\hline
\end{tabular}
}
\label{Table3}
\end{table*}

 \begin{table}[htbp]
\caption{Weight-based dropping performance: Ikura}
\centering
\resizebox{0.5\textwidth}{!}{%
\begin{tabular}{|c|c|c|c|}
\hline
\textbf{Target Weight(g)} & \textbf{Average 
dropped weight(g)} & \textbf{Accuracy(\%)} & \textbf{Standard Deviation}  \\
\hline
5 & 4.8 & 96\% & 0.42  \\
\hline
10 & 9.3 & 93\% & 0.95\\
\hline
15 & 15.4 & 97.4\% &1.07  \\
\hline
 20&20.8  & 96\% & 5.93 \\
\hline
\end{tabular}
}
\label{Table4}
\end{table}


\section{Perfromance Evaluation}
To evaluate the developed gripper's effectiveness, we conducted two core experiments and one demonstration using the right arm of a Nextage Open robot (Fig.5). Below are the details and results.

\subsection{Weight based dropping}    
 This experiment evaluates the gripper’s ability to handle challenging food items and accurately drop specific weights of them. The system operates in a closed-loop setup using a digital weighing scale placed beneath the target container. 
 
 The gripper's ability to drop food in a segregated and controlled approach combined with the weighing scale's feedback, allows the system to drop accurate weight of entangled, granualar, challenging food items.  We tested this on spaghetti (long, entangled, sticky) and ikura (granular, slippery), two items commonly found in Japanese bento boxes and known to be difficult for weight-based handling, and currently lack an effective weight-based solution. In the experiment, the gripper first picked up food from a large source bowl and then proceeded to drop the target weight into a designated bowl placed at the center of the workspace (Fig. 5). 


We tested four weight classes of spaghetti—10g, 30g, 50g, and 75g—based on typical quantities in Japanese food boxes, with ten trials per class. As the spaghetti’s entanglement and stickiness varied throughout the session, identical pickup conditions often resulted in different quantities. However, the gripper’s segregated pickup method enabled accurate, controlled dropping despite these variations. To evaluate robustness, the first five trials of each class were conducted early in the session, and the remaining five later, ensuring performance was tested under varying food states.\\
To evaluate the true effect of bits, we performed the same test on an ablated version of the gripper. The food bits were removed, and only the gripper with belt setup was used as the ablated gripper for comparison purposes. For the 10g class, the developed gripper had an average error of 4.18g, which is reasonable given each spaghetti strand weighs 3–6g. Accuracy improved with higher weight classes, yielding errors of 6.19g (30g), 9.5g (50g), and 5.8g (75g) (Table \ref{Table3}). In contrast, the ablated gripper failed to securely hold the spaghetti, typically picking only 10–12g regardless of depth or jaw width. Moreover, during the dropping phase, the ablated gripper was unable to regulate the release; the entire pickup load would fall at once, offering no fine-grained control. This resulted in significantly higher errors of 2.33g, 20.78g, 40.87g, and 54.21g for the 10g, 30g, 50g, and 75g target classes, respectively. These results highlight the critical role of food-specific bits in enabling secure pickup and precise, incremental dropping of challenging items, significantly enhancing the gripper’s interactive control and weight-based handling, even under difficult conditions.\\

For the ikura experiments, we used a gelatin-based ikura kit to prepare the required quantities. The bucket-shaped food bit, designed for slippery, granular, sticky, or semi-liquid foods (see Section 3.A.1), was used. For reference, each ikura ball weighs approximately 1g. We tested four weight classes—5g, 10g, 15g, and 20g—based on typical bento box usage\cite{slism_ikura}. The gripper achieved average errors under 1g across all classes (Table \ref{Table4}), corresponding to over 95\% accuracy, demonstrating the system’s high precision in handling such delicate and slippery food. This precision exceeds that for spaghetti due to the absence of extreme entanglement.
The ablated gripper failed to pick up ikura due to its slippery texture, so no comparison data could be recorded.; hence, we couldn't show the comparison data. Although a 95\% dropping accuracy over four weight classes strongly validates the effectiveness of both the bucket-shaped bits and the overall gripper design. \\

In conclusion, this experiment validates the effectiveness of the proposed gripper system and highlights the critical role of food-specific bits in enabling accurate pickup and weight-based dropping of challenging food items.




\subsection{Minimal weight drop ability}

A key advantage of the developed gripper is its ability to perform controlled, incremental releases of challenging food items, even when their physical properties—such as stickiness, entanglement, or slipperiness—vary significantly during handling. This capability enables the system to approach a target drop weight through a series of small, adaptive release steps, without requiring prior knowledge of the food's exact physical state.

To evaluate the gripper’s weight control capability, we tested two challenging food types: spaghetti (entangled, sticky) and ikura (granular, slippery). The gripper incrementally released food by rotating the servo motor in 20° steps, while measuring the dropped food weight to observe the granularity and consistency of control.

For spaghetti, the system demonstrated the ability to release as little as 3.47 g, corresponding to a single strand. Despite some natural variability due to entanglement and stickiness, the gripper exhibited a clear ability to release food in small steps across multiple rotations (Fig. 8).

For ikura, the smallest drop recorded was 2.42 g, consisting of approximately 2–3 individual roe balls (Fig. 8). Given ikura's slippery and granular nature, this again highlights the gripper’s precision and adaptability.

In contrast, a standard spaghetti tool—used as a baseline comparison—could only drop the entire grasped amount at once, offering no intermediate control. Most conventional food grippers exhibit similar behaviour, lacking the capacity for gradual release.

This experiment demonstrates that the developed gripper offers fine-grained, interactive control over the release of highly challenging food items. Unlike conventional grippers that drop the entire grasped portion at once, our system can consistently drop very small quantities, sometimes as low as a single strand of spaghetti or a few roe balls.  Such control is particularly advantageous in practical packaging tasks, where consistent food properties cannot be guaranteed, and real-time adaptation is necessary.


\begin{figure}[t]
\centering
\includegraphics[width=3.5in]{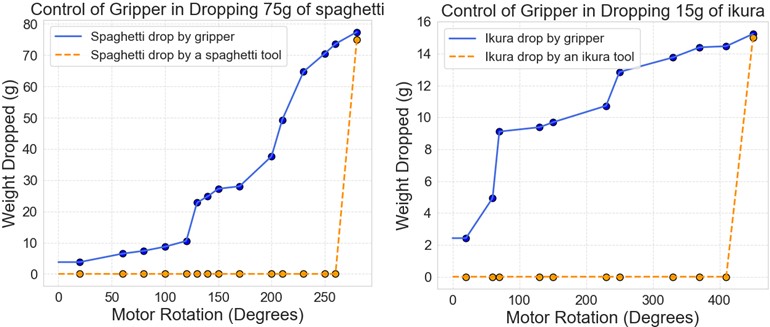}
\caption{Gripper's control over dropping small weights of challenging food items. Long entangled food item(left) and granular, sticky food item(right)}
\label{fig_1}
\end{figure}





\subsection{Belt assembly changing performance}
The system’s ability to replace the belt assembly and hence
the interacting food bit is vital to operate on a large range
of food items. The robustness of this system is of paramount
importance.


We evaluated the robustness of the passive belt-changing mechanism by conducting 10 consecutive switching trials, where the robotic arm executed a fixed point-to-point motion to replace the belt assembly. The system achieved a 100\% success rate across all trials, with no misalignments or mechanical failures. Each switching operation was completed in approximately 40 seconds, and the time remained consistent due to the deterministic nature of the movement. The switching process can be completed more quickly, but the duration was intentionally set to 40 seconds during testing to ensure safe and reliable operation.

The switching process uses no active actuators, relying solely on the robot’s precise motion and a passive design. This simplicity minimizes potential failure points and contributes to the system’s repeatability. The only requirement is that the robot maintains a positional accuracy within a 2 mm tolerance.

These results confirm the belt changer’s speed, reliability, and robustness. Its passive design and simple actuation make it ideal for frequent use in automated workflows needing rapid adaptation to varied food types.
\subsection{Food box packaging task}
To validate the food box packaging capability of the developed system, a demonstrative experiment was conducted(Supplementary Video). The robot was tasked with preparing two different food boxes: one with 50g spaghetti and 5g ikura; and another with 10g spaghetti and 20g ikura. The experiment evaluated the gripper’s ability to handle challenging food items, accurately drop target weights, and switch between different food-specific grippers. The system successfully completed the task with an overall accuracy of 91.25\%, and the total time taken was approximately 3.5 minutes(70 seconds per box and 40 seconds for tool changing). This scenario closely mirrors real-world bento box packaging. From these results, it can be concluded that the developed system is capable of automating bento packaging—exhibiting precise weight control, effective handling of challenging food items, and fast belt assembly changes for adapting to a wide range of food items.

\section{Conclusion}
This paper presents a hardware-centric approach to the challenge of automated bento box packaging. The proposed gripper features a triangular frame with an active belt surface onto which modular, food-specific elements called bits are attached. This design enables effective interaction with a variety of challenging food items, including deformable, slippery, and entangled ingredients. A key strength of the system is its precise control during food interaction, consistently dropping specific target weights, meeting the critical portion accuracy required in commercial food packaging. Additionally, the quick, passive belt assembly switching mechanism allows seamless transitions between food categories, accommodating the high variability of bento box contents.



Performance was evaluated through experiments: the gripper achieved over 80\% accuracy for spaghetti and over 95\% for ikura. It could release minimal amounts, down to a single strand of spaghetti or 2–3 ikura balls, enabling fine weight control through small adaptive steps. The belt-changing mechanism showed 100\% success, confirming its readiness to handle diverse food items.

The system ensures weight precision by releasing small portions, though this leads to an average serving time of 15 seconds per item. Future work will aim to increase speed without compromising precision. Apart from the technical aspect, several steps can be taken to ensure food safety and hygiene. During the ICRA 2024 Food Topping Challenge, food-contact areas were covered with food-safe tape to meet hygiene requirements. Other potential methods include applying FDA-compliant food-safe coatings (e.g. silicone, PTFE) or using detachable food-grade covers. Future work will focus on optimizing serving speed and improving accuracy for slippery or entangled foods.

\bibliographystyle{IEEEtran}
\bibliography{references}

\newpage

 




\vfill

\end{document}